\documentclass[letterpaper, 10pt, conference]{ieeeconf}
\IEEEoverridecommandlockouts  
\usepackage{algorithm}
\usepackage{amsmath}
\usepackage{amssymb}
\usepackage{bm}
\usepackage{color}
\usepackage{graphicx}
\usepackage[colorlinks, linkcolor=blue, urlcolor=blue]{hyperref}
\newcommand{\dd}[1]{\ensuremath {{\rm d}{#1}}}
\newcommand{\defeq}{\equiv}

\def\bfcd{\dot{\bfc}}

\def\bfxd{\dot{\bfx}}
\def\bfxid{\dot{\bfxi}}

\def\sd{\dot{s}}

\def\bfcdd{\ddot{\bfc}}

\def\bfxdd{\ddot{\bfx}}

\newcommand{\bfxi}{\boldsymbol{\xi}}

\newcommand{\bftau}{\boldsymbol{\tau}}

\newcommand{\bfc}{\ensuremath {\bm{c}}}

\newcommand{\bfe}{\ensuremath {\bm{e}}}
\newcommand{\bff}{\ensuremath {\bm{f}}}
\newcommand{\bfg}{\ensuremath {\bm{g}}}

\newcommand{\bfp}{\ensuremath {\bm{p}}}

\newcommand{\bfw}{\ensuremath {\bm{w}}}
\newcommand{\bfx}{\ensuremath {\bm{x}}}

\newcommand{\bfz}{\ensuremath {\bm{z}}}

\newcommand{\bfA}{\mathbf{A}}

\newcommand{\bfL}{\mathbf{L}}

\newcommand{\bfU}{\mathbf{U}}

\newcommand{\bfW}{\mathbf{W}}

\def\Ti{T_{\textit{i}}}
\def\com{\textit{com}}
\def\cop{\textit{cop}}
\def\dfz{\textit{$\delta$fz}}
\def\ileft{\textit{left}}
\def\iright{\textit{right}}
\def\kd{k_{\textit{d}}}
\def\ki{k_{\textit{i}}}
\def\kp{k_{\textit{p}}}
\def\kz{k_{\textit{z}}}
\def\qp{\textit{qp}}
\def\sd{\textit{d}}
\def\sm{\textit{m}}
\def\sz{\textit{0}}
\def\vdc{\textit{vdc}}

\def\RightFoot{\textit{ra}}
\def\RightFootCenter{\textit{rc}}
\def\LeftFoot{\textit{la}}
\def\LeftFootCenter{\textit{lc}}

\title{\LARGE \bf
    Stair Climbing Stabilization of the HRP-4 Humanoid Robot \\
    using Whole-body Admittance Control
}

\author{St\'ephane Caron, Abderrahmane Kheddar and Olivier Tempier%
    \thanks{The authors are with the Montpellier Laboratory of Informatics,
    Robotics and Microelectronics (LIRMM), CNRS--University of Montpellier,
    France. A. Kheddar is also with the CNRS--AIST Joint Robotics Laboratory
    (JRL), UMI3218/RL, Tsukuba, Japan. This work is supported in part by the
    H2020 EU project COMANOID \url{http://www.comanoid.eu/}, RIA No 645097.
    Corresponding author: {\tt\footnotesize stephane.caron@lirmm.fr}}%
}

\begin{document}

\maketitle
\thispagestyle{empty}
\pagestyle{empty}

\begin{abstract}
    We consider dynamic stair climbing with the HRP-4 humanoid robot
    as part of an Airbus manufacturing use-case demonstrator. We share
    experimental knowledge gathered so as to achieve this task, which HRP-4 had
    never been challenged to before. In particular, we extend walking
    stabilization based on linear inverted pendulum
    tracking~\cite{kajita2010iros} by quadratic programming-based wrench
    distribution and a whole-body admittance controller that applies both
    end-effector and CoM strategies. While existing stabilizers tend to use
    either one or the other, our experience suggests that the combination of
    these two approaches improves tracking performance. We demonstrate this
    solution in an on-site experiment where HRP-4 climbs an industrial
    staircase with 18.5~cm high steps, and release our walking controller as
    open source
    software.\footnote{\url{https://github.com/stephane-caron/lipm\_walking\_controller/}}
\end{abstract}

\section{Introduction}

Recently, humanoid robotics has reached a level of maturity that allows
considering deployments in large-scale manufacturing (\emph{e.g.}~aircraft and
shipyard), construction sites and nuclear power plants. These environments
are populated with stairs. In the case of aircraft manufacturing and shipyards, they allow
workers to travel between different shop-floor levels where assembly tasks are
required, which makes them a key challenge for any mobile robotics application.
Stair climbing was demonstrated as early as the first release of the Honda
humanoid robot in 1997~\cite{hirai1998icra}, yet it is still a challenging task
that humanoids rarely perform untethered. Robust stabilization is a critical
issue, not only to prevent robot falls, but for the safety of human co-workers
that are present and share the same working space\footnote{Performance and safety 
certification requirements are yet to be defined}.

\begin{figure}[t]
    \centering
    \vspace{1em}
    \includegraphics[width=0.6\columnwidth]{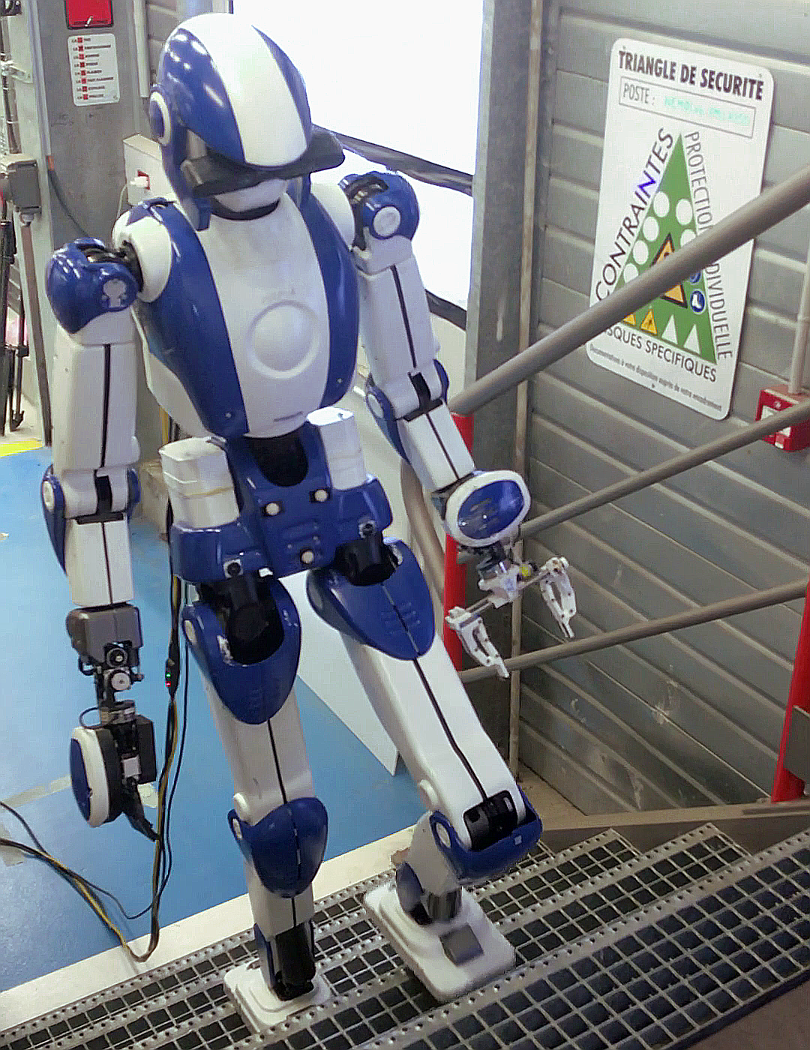}
    \caption{
        HRP-4 humanoid climbing an industrial staircase at the Airbus factory
        in Saint-Nazaire, France. Step height is 18.5~cm. The experiment was
        reproduced time and again over the course of two weeks spent on-site,
        and deemed robust enough to let the robot climb without safety ropes.
    }
    \label{fig:logo}
\end{figure}

In order to correct the deviation of their floating base from a reference
pattern, position-controlled robots adjust their contact forces with the
environment via \emph{admittance control}. To the exception of Honda humanoid
robots\footnote{Controllers reported by Honda include the \emph{model ZMP
    control} strategy~\cite{hirai1998icra, takenaka2009iros4} where saturation
    of ZMP constraints triggers recovery CoM accelerations and a corresponding
    update of the walking pattern~\cite{takenaka2009iros1, kamioka2018icra}.
    This integration of a switching control law with replanning behavior makes
    these controllers more advanced than linear feedback controllers.},
controllers found in the literature implement admittance either at the level of 
end-effectors~\cite{kajita2010iros, kim2009jirs, morisawa2012humanoids,
bouyarmane2019tro} or at the level of the CoM~\cite{nagasaka1999thesis, sugihara2002iros,
yokoi2004ijrr, englsberger2012humanoids}. Yet, these two
strategies are not mutually exclusive. In this work, we investigate a
whole-body admittance controller where both end-effector and CoM strategies are
applied simultaneously. Preliminary analysis in stair climbing simulations and
experiments suggests that a combination of these two approaches can improve
tracking performance.

\begin{figure*}
    \centering
    \includegraphics[width=17cm]{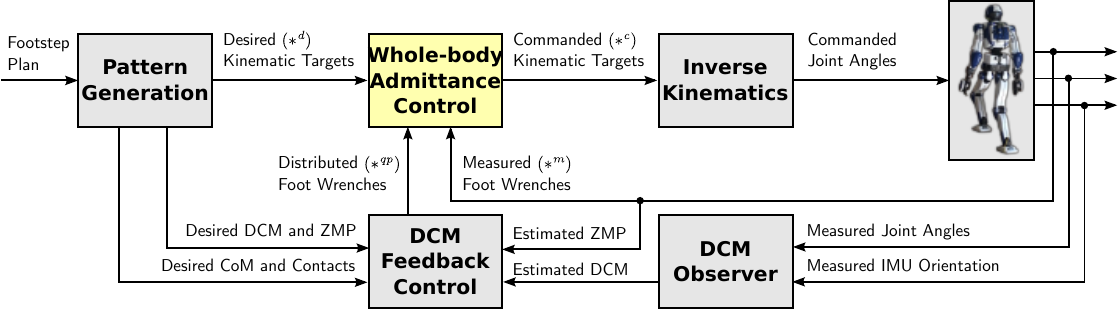}
    \caption{
        Overview of the walking and stair climbing controller for
        position-controlled robots based on feedback control of the divergent
        component of motion (DCM). The main contribution of this work lies in
        the combination of end-effector and CoM strategies to realize
        whole-body admittance control.
    }
    \label{fig:overview}
\end{figure*}

Figure~\ref{fig:overview} illustrates the components implemented in our walking
and stair climbing controller. The two main components for stabilization are:
\begin{itemize}
    \item \textit{DCM Feedback Control}
        (Section~\ref{sec:contact-wrench-control}), which computes desired
        contact wrenches to compensate deviation from the walking pattern. 
    \item \textit{Whole-body Admittance Control}
        (Section~\ref{sec:whole-body-admittance-control}), which allows a
        position-controlled robot to realize the desired contact wrenches.
\end{itemize}

We illustrate the performance of this controller in an on-site experiment in
the Airbus Saint-Nazaire site where the HRP-4 humanoid climbs a staircase with
18.5~cm steps. To the best of our knowledge, this is the first time that
dynamic stair climbing is demonstrated with HRP-4. The controller used in this
experiment is also open source and open to comments.$^1$

\section{DCM Feedback Control}
\label{sec:contact-wrench-control}

The goal of this first component is to regulate the robot's first-order
dynamics assuming control of its second-order dynamics. This amounts to decide
a \emph{net contact wrench} that compensates deviations from the walking
pattern.

Within the equations of motion of an articulated robot~\cite{wieber2005fast},
the centroidal dynamics is governed by the Newton-Euler equation:
\begin{equation}
    \label{eq:newton-euler}
    \begin{bmatrix} m \bfcdd \\ \dot{\bfL}_{\bfc} \end{bmatrix}
    =
    \begin{bmatrix} \bff \\ \bftau_{\bfc} \end{bmatrix}
    +
    \begin{bmatrix} m \bfg \\ \bm{0} \end{bmatrix}
\end{equation}
where $m$ denotes the total robot mass, $\bfg$ is the gravity vector, $\bfc$
the position of the center of mass (CoM) and $\bfL_{\bfc}$ the angular momentum
around $\bfc$. The net contact wrench $(\bff, \bftau_{\bfc})$ consists of the
resultant $\bff$ of external contact forces applied to the robot and their
moment $\bftau_{\bfc}$ around $\bfc$. The left-hand side of this equation
corresponds to the net motion of the robot, while the right-hand side
represents its interaction with the environment. The gist of locomotion is to
leverage these interaction forces to move the CoM (or similarly the translation
of the floating base) to a desired location.

\subsection{Linear Inverted Pendulum Mode}

A general walking pattern generator~\cite{carpentier2018tro} can provide both a
CoM trajectory $\bfc(t)$ and an angular-momentum trajectory $\bfL_{\bfc}(t)$.
Alternatively, this output can be reduced to the single CoM trajectory by
considering solutions where $\bfL_{\bfc} = \bm{0}$. The resulting model, known
as the Inverted Pendulum Mode (IPM)~\cite{pratt2007icra}, is expressive enough
for walking or stair climbing. It simplifies eq.~\eqref{eq:newton-euler}
to:
\begin{equation}
    \label{eq:ipm}
    \bfcdd = \lambda (\bfc - \bfz) + \bfg
\end{equation}
where the contact wrench is now characterized by a scaling factor $\lambda \geq
0$ and a \emph{zero-tilting moment point} (ZMP) $\bfz$.

Another working assumption acceptable for walking over a horizontal surface is
that of a constant CoM height $c_z = h$ above that surface. This gives rise to
the Linear Inverted Pendulum Mode~(LIPM)~\cite{kajita2001iros}:
\begin{equation}
    \label{eq:lipm}
    \bfcdd = \omega^2 (\bfc - \bfz)
\end{equation}
where $\omega = \sqrt{g / h}$, and we drop from now on the gravity vector by
considering only horizontal coordinates. The LIPM linearizes the
dynamics~\eqref{eq:ipm} of the IPM by turning the variable $\lambda$ into a
constant. Its contact wrench is characterized by the position $\bfz$ of the ZMP
on the contact surface.

These simplifications come at the expense of balance recovery strategies: the
IPM sacrifices the hip strategy~\cite{stephens2007humanoids} while the LIPM
sacrifices the height-variation strategy~\cite{koolen2016humanoids,
caron2018icra}. Accordingly, the dimension of the contact wrench decreases from
six to three in the IPM and two in the LIPM. A LIPM-based stabilizer such as
the one reported in the present paper can leverage these two force coordinates
to control two position coordinates, for instance the horizontal position of
the center of mass~\cite{kajita2010iros} or equivalently the rolling and
pitching angles of the floating base~\cite{takenaka2009iros4}.

\subsection{Feedback of the Divergent Component of Motion}

The \emph{divergent component of motion} (DCM) of the LIPM is defined by $\bfxi
= \bfc + {\bfcd}/{\omega}$. It allows a decomposition of the second-order
eq.~\eqref{eq:lipm} into two coupled first-order
systems~\cite{takenaka2009iros1}:
\begin{align}
    \bfxid & = \omega (\bfxi - \bfz) \label{eq:decoupled-1} \\
    \bfcd & = \omega (\bfxi - \bfc) \label{eq:decoupled-2}
\end{align}
While the DCM naturally diverges away from the ZMP by
eq.~\eqref{eq:decoupled-1}, eq.~\eqref{eq:decoupled-2} shows that the
CoM is guaranteed to converge to the DCM without being controlled. It is
therefore sufficient for locomotion to control only the DCM, rather than
\emph{e.g.} both the CoM position and velocity.

Walking pattern generation provides a trajectory $\bfc^\sd(t)$ in the linear
inverted pendulum mode~\eqref{eq:lipm}, from which one can derive
$\bfxi^\sd(t)$ and $\bfz^\sd(t)$. The DCM can be controlled around this
reference by proportional feedback~\cite{sugihara2009icra, englsberger2015tro,
wiedebach2016humanoids}:
\begin{equation}
    \label{eq:dcm-p}
    \bfxid = \bfxid^\sd + \kp (\bfxi^\sd - \bfxi^\sm)
\end{equation}
where $\kp$ is a positive feedback gain and the $*^\sm$ superscript denotes
estimated quantities.

An integral term can be added to eliminate steady-state
error~\cite{morisawa2012humanoids}, which would correspond here to an offset
between the CoM and ZMP positions when the robot is in static equilibrium:
\begin{equation}
    \label{eq:dcm-pi}
    \bfxid = \bfxid^\sd + \kp (\bfxi^\sd - \bfxi^\sm) + \ki \int (\bfxi^\sd - \bfxi^\sm)
\end{equation}
The integral term can include an anti-windup strategy such as saturation. We
used an exponential moving average, also known as leaky integrator:
\begin{equation}
    \int x = \frac{1}{\Ti} \int_{t=0}^T x(t) \exp\left({\frac{t - T}{\Ti}}\right) \dd{t}
\end{equation}
where $\Ti$ is the integrator time constant. This average does not wind up by
construction, and its implementation takes a single floating-point number in
memory.

Finally, a derivative term can be added to damp potential oscillations. From
eq.~\eqref{eq:decoupled-1}, it can be implemented indifferently by adding $\kd
(\bfxid^\sd - \bfxid^\sm)$ or substracting $\kz (\bfz^\sd - \bfz^\sm)$ to the
commanded $\bfxid$. We choose the latter in what follows.

From eq.~\eqref{eq:decoupled-1}, DCM feedback control can then be written in
terms of the commanded ZMP:
\begin{equation*}
    \bfz = \bfz^\sd - \left[1 + \frac{\kp}{\omega}\right] (\bfxi^\sd - \bfxi^\sm) - \frac{\ki}{\omega} \int (\bfxi^\sd - \bfxi^\sm) + \frac{\kz}{\omega} (\bfz^\sd - \bfz^\sm)
\end{equation*}
Note that the fact that the commanded $\bfz$ and measured $\bfz^\sm$ appear on
both sides of this equation comes from the unmodeled delay of admittance
control.\footnote{DCM feedback gains can be chosen by pole placement based on
an estimate of this delay~\cite{kajita2010iros, morisawa2012humanoids}.} This
commanded ZMP is equivalent to a commanded net contact wrench:
\begin{equation}
    \label{eq:wrench-pi}
    \begin{bmatrix} \bff \\ \bftau_{\bfc} \end{bmatrix}
    = m \begin{bmatrix} \omega^2 (\bfc - \bfz) - \bfg \\ \bm{0} \end{bmatrix}
\end{equation}
Thus, DCM feedback is a way to determine a net contact wrench that includes
both a feedforward term from the walking pattern and a feedback term to correct
CoM position and velocity deviations from their reference.



\subsection{Contact Wrench Distribution}
\label{sec:qp}

While stabilizers based on CoM admittance control~\cite{nagasaka1999thesis,
yokoi2004ijrr, englsberger2012humanoids, li2012icra} take the net wrench as
only input, those that include foot force control need to distribute this
wrench among contacts. This operation corresponds to the ZMP distributor of the
stabilizer by Kajita \emph{et al.}~\cite{kajita2010iros}. Meanwhile, the net
wrench obtained by DCM feedback is saturated in order to account for
feasibility constraints such as keeping the ZMP inside its support area. Both
distribution and saturation operations can be handled at once by formulating
the wrench distribution problem as a \emph{quadratic program}~(QP).

We will make use of spatial vector algebra~\cite{featherstone} to describe this
program. Let us denote by ${}^\sz\bfw_\textit{dcm}$ the net contact wrench
coordinates from eq.~\eqref{eq:wrench-pi} expressed in the inertial frame
$\sz$. Define ${}^\LeftFootCenter\bfw_\ileft$ and ${}^\LeftFoot\bfw_\ileft$ the
left foot contact wrench expressed respectively in the left sole center frame
$\LeftFootCenter$ and ankle frame $\LeftFoot$ (sole frame closest to the ankle
joint). Wrenches ${}^\RightFootCenter\bfw_\iright$ and
${}^\RightFoot\bfw_\iright$ are defined similarly.

\subsubsection{Constraints} in double support, the wrench distribution QP
consists of two constraints: contact stability, and a minimum pressure at each
contact:
\begin{align}
    \label{eq:qp-cons-1}
    \bfU {}^\LeftFootCenter\bfw_\ileft & \leq 0 &
    \bfU {}^\RightFootCenter\bfw_\iright & \leq 0 \\
    \label{eq:qp-cons-2}
    \bfe_{\textit{fz}} {}^\LeftFootCenter\bfw_\ileft & \geq p_\textit{min} &
    \bfe_{\textit{fz}} {}^\RightFootCenter\bfw_\iright & \geq p_\textit{min}
\end{align}
where $\bfU$ is the $16 \times 6$ matrix of the \emph{contact wrench
cone}~\cite{caron2015icra}. This matrix includes all three components of the
contact-stability condition: the Coulomb force friction cone,
center-of-pressure support area and net yaw moment boundaries. Meanwhile,
$\bfe_{\textit{fz}}$ is the basis vector that selects the resultant pressure of
a wrench, and $p_\textit{min}$ is a small threshold such as $15$~N. This
constraint avoids sending low-pressure targets to the foot force controller, as
fixed-gain admittance control tends to oscillate around contact switches for
such targets.

\subsubsection{Costs} the cost function of the wrench distribution QP weighs
three objectives:
\begin{align}
    \label{eq:qp-cost-1}
    & \left\| {}^\sz\bfw_\ileft + {}^\sz\bfw_\iright - {}^\sz\bfw_\textit{dcm} \right\|^2 \\
    \label{eq:qp-cost-2}
    & \left\| {}^\LeftFoot\bfw_\ileft \right\|^2_{\bfW_{\textit{ankle}}} +
    \left\| {}^\RightFoot\bfw_\iright \right\|^2_{\bfW_{\textit{ankle}}} \\
    \label{eq:qp-cost-3}
    & \left\| (1 - \rho) \bfe_{\textit{fz}} {}^\LeftFootCenter\bfw_\ileft - \rho \bfe_{\textit{fz}} {}^\RightFootCenter\bfw_\iright \right\|^2
\end{align}
First and foremost, the solution should realize the net
contact wrench as close as possible~\eqref{eq:qp-cost-1}. Second, it should
minimize ankles torques~\eqref{eq:qp-cost-2}, where $\bfW_{\textit{ankle}}$ is
a diagonal weight matrix with $1$ for ankle torques and a small value
$\epsilon$ for all other components. Finally,
the pressure ratio should be as close as possible~\eqref{eq:qp-cost-3} to a
prescribed value $\rho$. This last term regularizes the discontinuity in force
output that occurs in acceleration-based whole-body controllers when adding or
removing contacts~\cite{saab2013tro, bouyarmane2019tro}. The prescribed
pressure ratio ranges from $\rho_{\textit{init}} \in \{0, 1\}$ at the beginning
of the double support phase to $1 - \rho_{\textit{init}}$ at the end of it.

Although we presented and implemented it as a quadratic program, this
optimization is in essence a lexicographic optimization~\cite{escande2014ijrr}
whose four levels are~\eqref{eq:qp-cons-1}--\eqref{eq:qp-cons-2},
\eqref{eq:qp-cost-1}, \eqref{eq:qp-cost-2} and \eqref{eq:qp-cost-3}. We
approximate this behavior by setting cost weights to $10000$ for
\eqref{eq:qp-cost-1}, $100$ for \eqref{eq:qp-cost-2} and $1$ for
\eqref{eq:qp-cost-3}. Note that the latter two costs are omitted during single
support where there is no force redundancy and the net-wrench
cost~\eqref{eq:qp-cost-1} is enough to define a single optimum.

\section{Whole-body admittance control}
\label{sec:whole-body-admittance-control}

Whole-body admittance control implements feedback control of the desired force
targets issued by DCM feedback and wrench distribution, while otherwise
following the position targets prescribed by the walking pattern.

\subsection{Foot damping control}

Admittance control applied at the ankle joint has been referred to as ground
reaction force control~\cite{hirai1998icra, takenaka2009iros4}, foot damping
control~\cite{kajita2010iros}, or foot adjusting control~\cite{yokoi2004ijrr}.
It implements the first stabilization strategy from Section 4.5.1 of the
\emph{Introduction to Humanoid Robotics}~\cite{kajita}.

Let us denote by $(\theta^\textit{c}_r, \theta^\textit{c}_p)$ the
\emph{commanded} (see Figure~\ref{fig:overview}) roll and pitch angles of the
ankle joint of a foot in contact with the environment. We apply the following
damping control law\footnote{Damping control is a shorthand for first-order
admittance control.} to track a desired CoP:
\begin{align}
    \label{eq:foot-damping-control}
    \begin{bmatrix} \dot{\theta}^c_r \\ \dot{\theta}^c_p \end{bmatrix}
    &
    = \bfA_\cop (\bfp^\qp \times \bff^\sm - \bftau^\sm)
    \\
    \bfA_\cop
    & \defeq \begin{bmatrix}
        A_{\cop,y} & 0 & 0 \\
        0 & A_{\cop,x} & 0
    \end{bmatrix}
\end{align}
where $\bfp^\qp = [p^\qp_x\ p^\qp_y\ 0]$ denotes the target CoP position in the
foot frame provided by the wrench distribution QP, and $(\bff^\sm, \bftau^\sm)$
is the measured contact wrench expressed at the origin of the foot frame. The
matrix $\bfA_\cop$ of admittance gains $(A_{\cop,x}, A_{\cop,y})$ is used to tune
the responsiveness of the task: a higher $A_{\cop,y}$ implies that the foot
will roll faster in reaction to lateral CoP deviations, and similarly a higher
$A_{\cop,x}$ implies that the foot will pitch faster in reaction to sagittal
CoP deviations. We further clamped the absolute values of the resulting
velocities $(\dot{\theta}^c_r, \dot{\theta}^c_p)$ to $0.2$ rad/s.

The above eq.~\eqref{eq:foot-damping-control} is adapted
from~\cite{kajita2010iros}, with the slight difference that we track the
desired CoP rather than a desired torque. The two approaches are equivalent
under accurate foot pressure difference tracking, but in situations where the
latter is degraded, the CoP formulation naturally defines the
pressure-dependent admittance coefficients identified
in~\cite{pajon2017humanoids}. This task can be extended to include integral and
derivative terms of the measured wrench~\cite{bouyarmane2019tro}. It may also
be improved by a model of the flexibility located between the ankle joint and
foot sole~\cite{takenaka2009iros4, kajita2001icra}, which we do not include.

\subsection{Foot force difference control}

In a walking gait, double support phases are used to transfer the net ZMP from
one support foot to the next. It is therefore helpful to servo not only the CoP
targets provided at each foot, but also their pressure. For this purpose,
Kajita \emph{et al.}~\cite{kajita2010iros, kajita2013icra} introduced
\textit{foot force difference control} (FFDC). Denoting by $(v_{Lz}, v_{Rz})$
the respective velocities of the left and right foot in their sole frames, FFDC
can be implemented as:
\begin{align}
    \label{eq:ffdc-1} v^\textit{c}_{Lz} & = v_{Lz}^\sd - 0.5 v_\dfz + 0.5 v_\vdc \\
    \label{eq:ffdc-2} v^\textit{c}_{Rz} & = v_{Rz}^\sd + 0.5 v_\dfz + 0.5 v_\vdc \\
    \label{eq:ffdc-3} v_\dfz & \defeq A_\dfz \left[(f_{Lz}^\qp - f_{Rz}^\qp) - (f_{Lz}^m - f_{Rz}^m)\right] \\
    \label{eq:ffdc-4} v_\vdc & \defeq T_\vdc^{-1} \left[(p_{Lz}^d + p_{Rz}^d) - (p_{Lz}^c + p_{Rz}^c)\right]
\end{align}
The velocity term $v_\dfz$ implements a damping control that lifts the foot in
excess of pressure and lowers the other one. It is tuned by the admittance gain
$A_\dfz$. The second velocity term $v_\vdc$ is added for vertical drift
compensation. It retrieves the same average foot altitude as in the walking
pattern, tuned by a frequency gain $T_\vdc^{-1}$ set to $1$~Hz in practice.
This choice of a velocity formulation~\eqref{eq:ffdc-1}--\eqref{eq:ffdc-4} of
FFDC rather than the position one from~\cite{kajita2010iros} is contingent to
our inverse kinematics and yields the same behavior.

An implicit side effect of FFDC is that it \emph{increases CoM compliance}. To
illustrate this remark, consider the example of a constant external push
applied to laterally: with only foot damping control, the robot will resist it
by tilting its feet, while with FFDC it will lift the leg opposite to the push,
resulting (as gravity maintains contact) in a CoM displacement toward that leg.
As such, we may venture to say that our reference
controller~\cite{kajita2010iros} implicitly included a form of CoM admittance
control.

\subsection{CoM Admittance Control}

Admittance control applied at the CoM has been referred to as ZMP compliance
control~\cite{nagasaka1999thesis}, ZMP damping control~\cite{yokoi2004ijrr},
position-based ZMP control~\cite{englsberger2012humanoids} or horizontal
compliance control~\cite{li2012icra}. It implements the third stabilization
strategy from Section 4.5.1 of the \emph{Introduction to Humanoid
Robotics}~\cite{kajita}, and should not be confused with the model ZMP control
(fifth strategy) applied on Honda robots~\cite{hirai1998icra,
takenaka2009iros4, kamioka2018icra}. The former adds CoM accelerations at all
times, the latter only upon saturation of a ZMP constraint.

We apply the following admittance control law:
\begin{align}
    \label{eq:com-admittance}
    \bfcdd^\textit{c} & = \bfcdd^\sd + \bfA_\com (\bfz^\sm - \bfz^\qp) \\
    \bfA_\com & \defeq \begin{bmatrix}
        A_{\com,x} & 0 & 0 \\
        0 & A_{\com,y} & 0
    \end{bmatrix}
\end{align}
where $\bfcdd^\sd$ is the feedforward CoM acceleration from the walking
pattern, $\bfz^\sm$ is the ZMP of the measured net contact wrench and
$\bfz^\qp$ is the ZMP of the net contact wrench output by the distribution QP
(Section~\ref{sec:qp}). The matrix $\bfA_\com$ of admittance gains
$(A_{\com,x}, A_{\com,y})$ is used to tune the responsiveness of the task: the
higher the gain, the faster the CoM will accelerate toward the measured ZMP so
as to move it back towards the desired one.

\begin{figure}[t]
    \centering
    \includegraphics[width=\columnwidth]{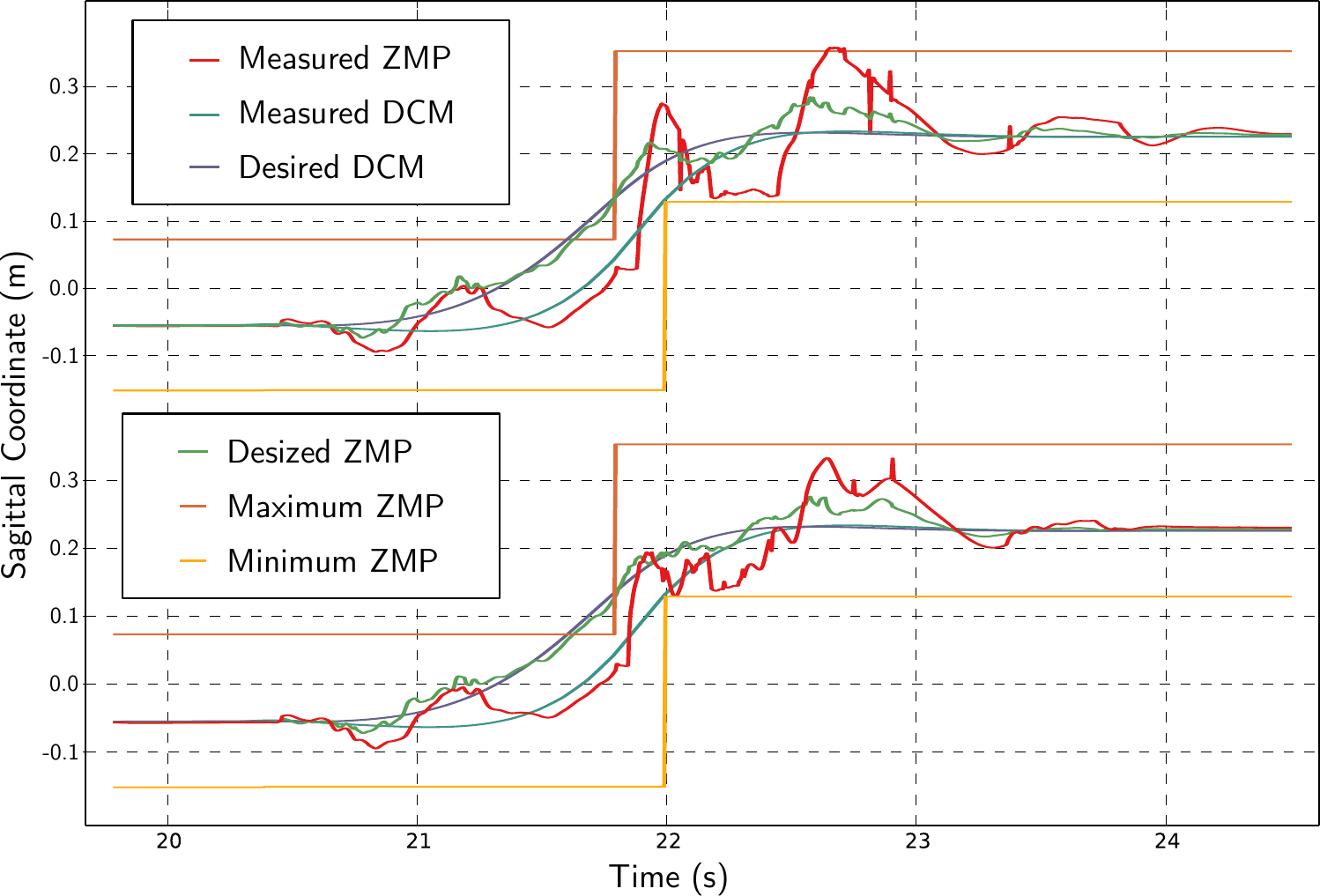}
    \caption{
        Effect of applying CoM admittance control in combination with foot
        damping control. \textbf{Top:} $A_{\com,x}=0$ and $A_{\com,y}=0$.
        \textbf{Bottom:} $A_{\com,x}=20$ and $A_{\com,y}=10$. In this
        simulation, the robot steps on an 18.5~cm step but tilts back on its
        heel during left-foot support. CoM admittance control (bottom) helps
        mitigate this effect.
    }
    \label{fig:com-admittance}
\end{figure}

Figure~\ref{fig:com-admittance} shows a simulation example of step climbing
without and with CoM admittance control, both foot damping control laws being
active. The CoM admittance law does not seem to conflict with the end-effector
ones. On the contrary, it improves both DCM and ZMP tracking noticeably.

\subsection{Inverse Kinematics}

Commanded velocities and accelerations are sent to a weighted task-based
inverse kinematics solver~\cite{bouyarmane2019tro, saab2013tro}. The following
tasks are considered simultaneously:
\begin{itemize}
    \item Maintain foot contact(s) (weight: 10000)
    \item CoM position and velocity tracking (weight: 1000)
    \item Swing foot position and velocity tracking (weight: 500)
    \item Bend the chest to a prescribed angle (weight: 100)
    \item Keep the pelvis upright (weight: 10)
    \item Regularizing half-sitting joint configuration (weight: 10)
\end{itemize}
Each task implements an acceleration-based tracking law:
\begin{align}
    \ddot{\bfx} & = K (\bfx^c - \bfx) + B (\bfxd^c - \bfxd) + \bfxdd^c
\end{align}
Task damping coefficients $B$ are set by default to their critical value $2
\sqrt{K}$, with the exception of foot contact tasks where we use $B=300$~Hz and
$K=1$~Hz$^2$. In single support where foot force difference control is
disabled, the translation stiffness of the support foot task is increased to
$K=1000$~Hz$^2$ for vertical drift compensation. 

For a foot $F \in \{L, R\}$ in contact, the target velocity $\bfxd^c_F$ of the corresponding foot
contact task is defined from eq.~\eqref{eq:foot-damping-control} and
\eqref{eq:ffdc-1}--\eqref{eq:ffdc-2}.
At present, we kept $\bfxdd^c_F = \bm{0}$ and $\bfx^c_F$ fixed to the desired
contact location. For an improved behavior, the latter can be updated by an
integral of the $\bfxd^c_F$ so that all targets $\bfx^c_F, \bfxd^c_F,
\bfxdd^c_F$ become consistent~\cite{bouyarmane2019tro}.

\section{Experiments}

\begin{figure}[t]
    \centering
    \includegraphics[width=\columnwidth]{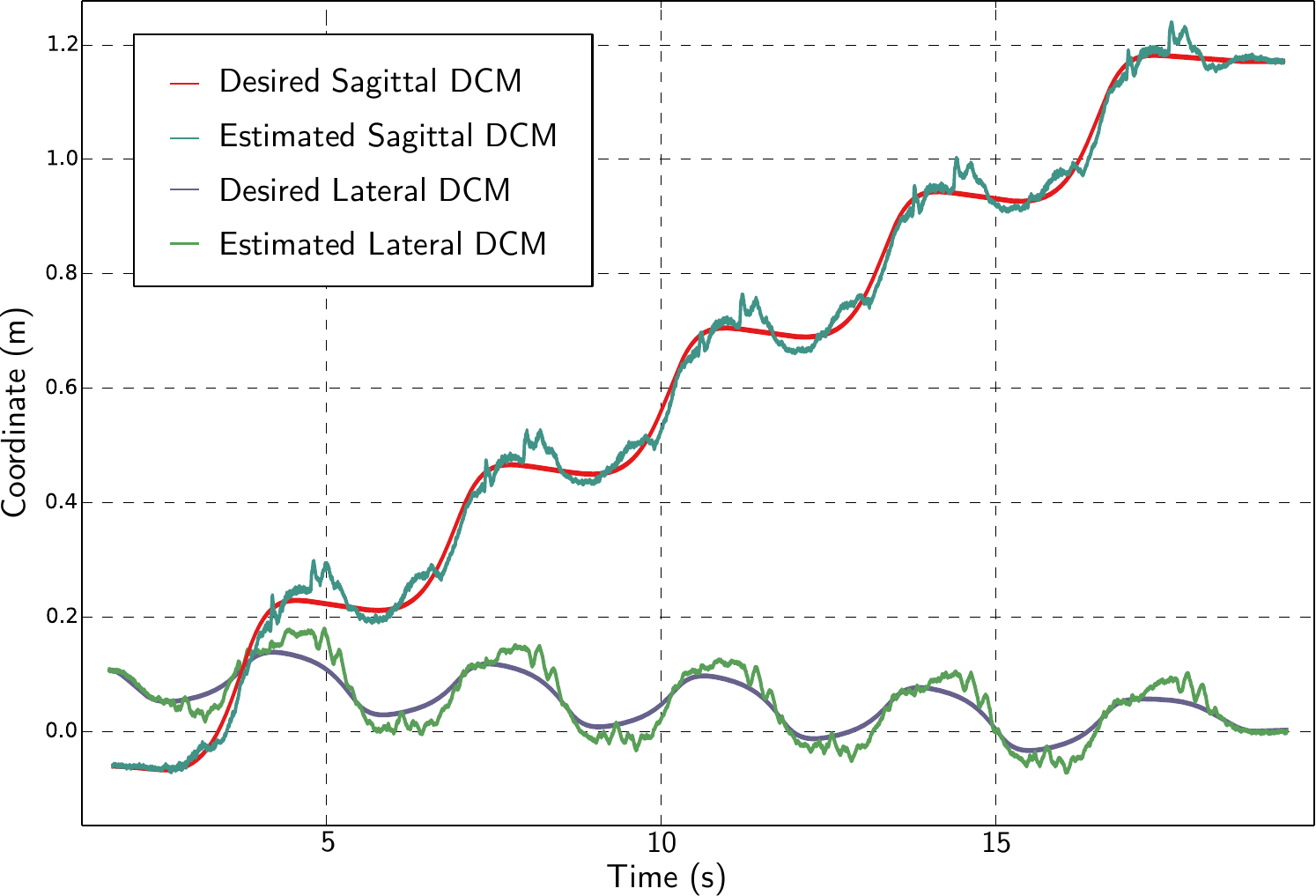}
    \caption{
        DCM tracking performance while climbing the factory staircase with
        18.5~cm steps. Swing leg motions are not accounted for in the walking
        pattern and drive the DCM away from its reference at each step. These
        disturbances are compensated by the stabilizer. (Note: trajectories
        slightly slanted as the staircase was not exactly aligned with our
        inertial frame.)
    }
    \label{fig:dcm-tracking}
\end{figure}

We implemented our controller in the \emph{mc\_rtc} framework (see the Appendix
for details on other components) and carried out experiments with the HRP-4
humanoid robot~\cite{kaneko2011iros}.

\subsection{Demonstration environment}

Experiments were carried out on-site at the Airbus factory located in
Saint-Nazaire, France. In the final demonstration, the robot walked to the
staircase, climbed it, walked to a designated area inside the fuselage of an
A350 aircraft and performed an assembly task before walking out. The staircase
climbed to access the assembly area had five steps of length $24$~cm and height
$18.5$~cm.

\begin{figure}[t]
    \centering
    \includegraphics[width=\columnwidth]{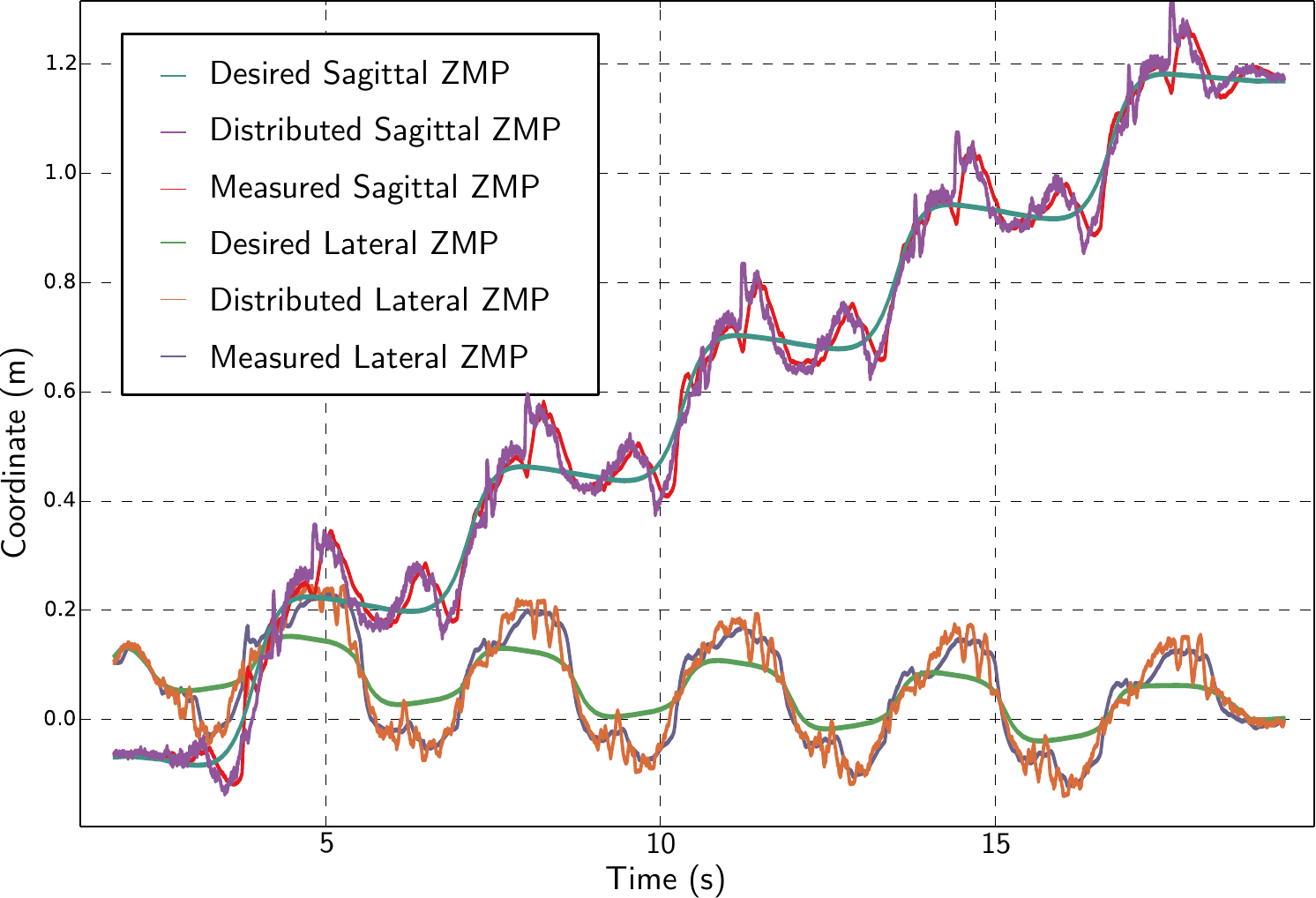}
    \caption{
        ZMP tracking performance while climbing the factory staircase with
        18.5~cm steps. The distributed ZMP is driven away from the walking
        pattern reference to compensate for DCM errors. Whole-body admittance
        control regulates the ZMP to the distributed one. (Note: trajectories
        slightly slanted as the staircase was not exactly aligned with our
        inertial frame.)
    }
    \label{fig:zmp-tracking}
\end{figure}

In preliminary experiments with varying step heights, shown in the accompanying
video,\footnote{\url{https://www.youtube.com/watch?v=vFCFKAunsYM}} we also used
a cable-driven parallel robot to act as safety crane for HRP-4. This robot was
developed in our laboratory and consists of eight actuators, four of which were
used. Cables were attached to a taylored holder connecting safety ropes to the
shoulders of the humanoid. The system was remote-controlled by a human
operator, making sure that the ropes stay loose while trying to avoid hitting
the robot as it climbs (bonus robustness checks otherwise).

\subsection{Results}

We confirmed that HRP-4 can dynamically climb the industrial staircase, as
shown in Figure~\ref{fig:logo} and in the accompanying video. The performance
was reproduced time and again over the course of two weeks spent on-site at the
Airbus site, and deemed robust enough to let the robot climb without safety
ropes in the latter experiments. The robot climbs the stairs in 18~s with 1.4~s
single-support and 0.2~s double-support durations. DCM feedback was tuned as
follows:
\begin{center}
    \begin{tabular}{|c|c|c|c|c|}
        \hline
        $\kp$ [Hz] & $\ki$ [Hz] & $\kz$ [Hz] & $\Ti$ [s] \\
        \hline
        5 & 20 & 2 & 20 \\
        \hline
    \end{tabular}
\end{center}
Meanwhile, admittance control parameters were set to the following values (in
[s.kg$^{-1}$] and [kg$^{-1}$] respectively):
\begin{center}
    \begin{tabular}{|c|c|c|c|c|}
        \hline
        $A_{\cop,x}$ & $A_{\cop,y}$ & $A_{\dfz}$ & $A_{\com,x}$ & $A_{\com,y}$ \\
        \hline
        0.01 & 0.01 & 0.0001 & 20 & 10 \\
        \hline
    \end{tabular}
\end{center}
DCM and ZMP tracking performance are reported in Figure~\ref{fig:dcm-tracking}
and~\ref{fig:zmp-tracking} respectively.

The overall run time of a controller cycle on a consumer laptop computer is
around $1.0 \pm 0.4$~ms, which fits within the $5$~ms of HRP-4's control loop.
Most of this time is spent solving the inverse kinematics QP ($0.4 \pm 0.1$~ms)
and the wrench distribution QP ($0.3 \pm 0.1$~ms during double support) using
the LSSOL least-squares solver for both. Running at a lower frequency, the
model predictive control QP is solved in $0.3 \pm 0.1$~ms using the QLD solver.

To avoid collisions, the apex of swing foot trajectories for each step is set
to 24~cm, which is a first source of DCM disturbance. The second main cause are
the CoM height variations at each step that disturb the horizontal ZMP
backwards. We mitigated this by delaying CoM lift to the end of the step,
unfortunately thus increasing knee torques as well. A better way to improve
this in future work will be to switch to a pattern generation method taking
height variations into account~\cite{caron2018icra, englsberger2015tro,
brasseur2015humanoids}.

\subsection{Practicalities}

One of the most precious tools at our disposal during our trials and errors was
the Choreonoid environment and its dynamics simulator~\cite{nakaoka2007rsj}, in
which we could reproduce most of the phenomena encountered in practice. The
ability to test controllers in fast simulations rather than slow experiments is
a serious enabler, and for humanoid robotics, Choreonoid outperformed
alternatives like V-REP or Gazebo in terms of both realism and real-time
performance.

During our first experiments, the robot would systematically servo-off during
the second (most knee-torque intensive) swing phase of step climbing. This was
caused by a drop of voltage due to a maximum current setting of 5~A on the
power supply. We increased this threshold and observed peak current draws
reaching up to 13~A. We estimate the peak power consumption to be around 750~W.
For comparison, ASIMO consumes 600--900 W when its servomotors are turned on,
and around 1000~W during stair climbing~\cite{hirose2007}. The consumption gap
between the two robots is mostly owed to the design of
HRP-4~\cite{kaneko2011iros}, which is both lighter (40~kg versus 50~kg) and
taller than its Honda sibling (1.5~m versus 1.3~m), allowing it to bend its
knees less while climbing.

Our initial plan was to climb (\textit{i}) a single step, then (\textit{ii})
the staircase with double-support phases at each step, and finally
(\textit{iii}) a more human-like stair climbing with exactly one foot contact
per step. We presently report on (\textit{ii}) but not (\textit{iii}), as a
mechanical transmission issue prevents our robot from performing with its right
leg the motions that it achieves with the left one, even for lower step
heights.

\section{Related stair climbing works}

Stair climbing for bipeds with ZMP-based stabilizers started as early as 1993,
when the Honda E6 prototype climbed staircases thanks to the stabilization
strategies developed by Takenaka~\cite{hirose2007}. This method was showcased
in 1997 for the public release of the P2 humanoid robot~\cite{hirai1998icra}.
Stair climbing was also demonstrated in 2002 on the prototype HRP-1S of the HRP
series~\cite{yokoi2004ijrr}.

However, the stabilizer component provided with robots of the HRP series is
mainly designed for walking on overall level ground. Climbing over small
staircases with 10~cm steps has been reported on HRP-2~\cite{carpentier2018tro,
michel2007iros}, walking with bent knees to avoid undesired behavior close to
the knee kinematic singularity. In~\cite{kim2009jirs}, KHR-2 climbed stairs
with 12~cm steps. In~\cite{carpentier2018tro}, HRP-2 also climbed stairs with
15~cm steps while grabbing a handrail.

Step heights above 20~cm have been demonstrated, yet with slower gaits.
In~\cite{zhang2017ijhr}, HRP-4 climbed a 24~cm step, but without stabilization
and with a quasi-static motion lasting more than 80~s. During the DARPA
Robotics Challenge, six teams successfully climbed a staircase with four 23~cm
steps, yet with slow motions and doing frequent pauses as a result of the
challenge's conditions. Shank collisions also become a concerning problem for
higher step heights. To deal with this issue, team KAIST climbed stairs
backwards~\cite{lim2016jfr} while teams IHMC and ESCHER used partial
footholds~\cite{hopkins2015humanoids, wiedebach2016humanoids}.

\section{Conclusion}

In this paper, we reported on the state-of-the-art of walking stabilization
by DCM feedback control and suggested two improvements: a wrench distribution
quadratic program, and a whole-body admittance controller combining both
end-effector and CoM strategies. We applied the resulting controller in a
dynamic stair climbing experiment over 18.5~cm steps, performed in an
industrial environment at the Airbus Saint-Nazaire site.

Our stabilizer has a number of gains to tune, some of which interact with each
other. For instance, lowering foot CoP admittances allows one to raise the DCM
feedback gain $\kp$ to larger values before reaching the unstable regime. The
elephant in the room hindering our understanding here is the unmodeled
flexibility below foot ankles, modeled as a first-order ZMP delay
in~\cite{kajita2010iros, morisawa2012humanoids}. Future work will require us to
investigate this question, and at least another one: how to prevent or mitigate
touchdown impacts?

\section*{Acknowledgments}

The authors warmly thank Kevin Chappellet for his help with robot hardware,
Pierre Gergondet for developing \emph{mc\_rtc}, Daniele De Simone and Arnaud
Tanguy for their assistance with experiments, Tomomichi Sugihara and
Pierre-Brice Wieber for helpful and friendly discussions, as well as Junhyeok
Ahn, Andrea Del Prete and Louise Scherrer for their feedback. The cable-driven
parallel robot used during preliminary experiments was realized thanks to the
support of the European Union through FEDER Grant No 49793.

\appendix

\subsection{Walking Pattern Generation}

We generate walking patterns by linear model predictive
control~\cite{wieber2006humanoids} over pre-defined footstep locations. The
corresponding QP minimizes three weighted costs:
\begin{itemize}
    \item ZMP deviation from a reference (weight: 1000)
    \item CoM velocity deviation from a reference (weight: 10)
    \item CoM jerk (weight: 1)
\end{itemize}
The reference ZMP trajectory $\bfz_\textit{ideal}(t)$ consists of straight
lines connecting foot ankle frames. The QP also enforces the following three
constraints:
\begin{itemize}
    \item Feasibility: ZMPs lie in their support polygons
    \item Terminal constraint 1: the ZMP ends on $\bfz_\textit{ideal}(T)$ 
    \item Terminal constraint 2: the DCM ends on $\bfz_\textit{ideal}(T)$ 
\end{itemize}
where $T = 1.6$~s is the duration of the predictive horizon and the sampling
period is set to $100$~ms. We let predictive control update the CoM reference
$\bfc^\sd(t)$ and its derivatives by open-loop integration rather than
a closed-loop approach~\cite{villa2017humanoids}. 

\subsection{DCM observer}

Although we plan to evaluate methods that take into account foot
flexibilities~\cite{benallegue2017humanoids, flayols2017humanoids}, for now we
use a simple kinematics estimator based on Kalman filtering to estimate the
orientation of the floating-base and an anchor-point assumption to estimate its
translation. The CoM position is then derived by forward kinematics of
joint-encoder readings, and its velocity by low-pass filtering.






\bibliographystyle{IEEEtran}
\bibliography{refs}

\end{document}